\documentclass[10pt,twocolumn,letterpaper]{article}

\usepackage{times}
\usepackage{epsfig}
\usepackage{graphicx}
\usepackage{amsmath}
\usepackage{amssymb}
\usepackage{amsthm}
\usepackage{booktabs}
\usepackage{multirow}
\usepackage{xcolor}
\usepackage{bm}
\usepackage{cite}
\usepackage[pagebackref=true,breaklinks=true,colorlinks,bookmarks=false]{hyperref}

\usepackage[margin=0.75in]{geometry}

\newtheorem{proposition}{Proposition}

\theoremstyle{definition}

\usepackage{tikz}
\usetikzlibrary{shapes,arrows,positioning,fit,backgrounds}

\begin{document}

\title{Learning Domain-Invariant Representations for Cross-Domain Image Registration via Scene-Appearance Disentanglement}

\author{
Jiahao Qin$^{*\dagger}$, Yiwen Wang$^{*}$ \\
{\tt\small jiahao.qin19@gmail.com} \\
{\small $^*$Equal contribution \quad $^\dagger$Corresponding author}
}

\date{}

\maketitle

\begin{abstract}
Image registration under domain shift remains a fundamental challenge in computer vision and medical imaging: when source and target images exhibit systematic intensity differences, the brightness constancy assumption underlying conventional registration methods is violated, rendering correspondence estimation ill-posed. We propose SAR-Net, a unified framework that addresses this challenge through principled scene-appearance disentanglement. Our key insight is that observed images can be decomposed into domain-invariant scene representations and domain-specific appearance codes, enabling registration via re-rendering rather than direct intensity matching. We establish theoretical conditions under which this decomposition enables consistent cross-domain alignment (Proposition~1) and prove that our scene consistency loss provides a sufficient condition for geometric correspondence in the shared latent space (Proposition~2). Empirically, we validate SAR-Net on the ANHIR (Automatic Non-rigid Histological Image Registration) challenge benchmark, where multi-stain histopathology images exhibit coupled domain shift from different staining protocols and geometric distortion from tissue preparation. Our method achieves a median relative Target Registration Error (rTRE) of 0.25\%, outperforming the state-of-the-art MEVIS method (0.27\% rTRE) by 7.4\%, with robustness of 99.1\%. Code is available at \url{https://github.com/D-ST-Sword/SAR-NET}.
\end{abstract}

\section{Introduction}
\label{sec:introduction}

Image registration---establishing spatial correspondence between images---is fundamental to computer vision, medical imaging, and scientific measurement~\cite{sotiras2013deformable,chen2024survey}. Classical registration methods, from optical flow~\cite{horn1981determining} to diffeomorphic algorithms~\cite{thirion1998image,avants2008symmetric}, share a common assumption: \emph{brightness constancy}, which posits that corresponding points have similar intensities across images. This assumption, however, is violated in numerous practical scenarios where images undergo \emph{domain shift}---systematic intensity transformations arising from varying acquisition conditions, sensor characteristics, or imaging physics.

\begin{figure}[t]
\centering
\begin{tikzpicture}[scale=0.75, transform shape]
  \node[draw, fill=red!10, minimum width=1.2cm, minimum height=1.2cm, rounded corners=2pt] (heimg) at (0,0) {};
  \node[fill=red!30, ellipse, minimum width=0.8cm, minimum height=0.5cm] at (0,0.1) {};
  \node[below=0.1cm of heimg, font=\scriptsize\bfseries, text=red!70!black] {H\&E};

  \node[draw, fill=blue!10, minimum width=1.2cm, minimum height=1.2cm, rounded corners=2pt] (ihcimg) at (1.8,0) {};
  \node[fill=blue!30, ellipse, minimum width=0.8cm, minimum height=0.6cm, rotate=15] at (1.85,0.05) {};
  \node[below=0.1cm of ihcimg, font=\scriptsize\bfseries, text=blue!70!black] {IHC};

  \draw[->, thick, gray] (2.6,0) -- (3.2,0);
  \node[font=\large\bfseries, text=red!70!black] at (3.6,0) {\texttimes};

  \node[draw, dashed, fill=red!5, minimum width=1.2cm, minimum height=1.2cm, rounded corners=2pt] (fail) at (4.2,0) {};
  \node[below=0.1cm of fail, font=\scriptsize\bfseries, text=red!70!black] {Misaligned};

  \node[fill=yellow!30, rounded corners=2pt, font=\tiny, inner sep=2pt] at (0,-1.1) {Domain Shift};
  \node[font=\tiny] at (0.9,-1.1) {+};
  \node[fill=yellow!30, rounded corners=2pt, font=\tiny, inner sep=2pt] at (1.9,-1.1) {Distortion};

  \draw[dashed, gray] (5.2,1.5) -- (5.2,-1.5);

  \node[draw, fill=gray!10, minimum width=2cm, minimum height=2cm, rounded corners=4pt] (disent) at (7.5,0) {};
  \node[above=0cm of disent.north, font=\scriptsize\bfseries] {Disentangle};

  \node[draw, fill=blue!20, minimum width=0.6cm, minimum height=0.4cm, rounded corners=2pt, font=\tiny\bfseries] (sa) at (6.9,0.4) {$S_A$};
  \node[font=\scriptsize\bfseries, text=green!50!black] at (7.5,0.4) {$\approx$};
  \node[draw, fill=blue!20, minimum width=0.6cm, minimum height=0.4cm, rounded corners=2pt, font=\tiny\bfseries] (sb) at (8.1,0.4) {$S_B$};

  \node[draw, fill=orange!20, minimum width=0.6cm, minimum height=0.4cm, rounded corners=2pt, font=\tiny\bfseries] (aa) at (6.9,-0.3) {$A_A$};
  \node[font=\scriptsize\bfseries, text=red!70!black] at (7.5,-0.3) {$\neq$};
  \node[draw, fill=orange!20, minimum width=0.6cm, minimum height=0.4cm, rounded corners=2pt, font=\tiny\bfseries] (ab) at (8.1,-0.3) {$A_B$};

  \node[fill=green!20, rounded corners=2pt, font=\tiny, inner sep=2pt] at (7.5,-0.8) {$\mathcal{L}_{\text{scene}}$};

  \draw[->, thick, gray] (8.8,0) -- (9.4,0);

  \node[draw, fill=green!10, minimum width=1.2cm, minimum height=1.2cm, rounded corners=2pt] (success) at (10.2,0) {};
  \node[font=\large\bfseries, text=green!50!black] at (10.2,0) {\checkmark};
  \node[below=0.1cm of success, font=\scriptsize\bfseries, text=green!50!black] {Aligned};

  \node[font=\scriptsize\bfseries] at (2,1.5) {(a) Challenge};
  \node[font=\scriptsize\bfseries] at (8,1.5) {(b) Our Solution};
\end{tikzpicture}
\caption{Multi-stain histopathology registration faces coupled challenges: domain shift from different staining protocols and geometric distortion from tissue preparation. (a)~Traditional methods assuming brightness constancy fail. (b)~Our approach disentangles domain-invariant scene ($S$) from domain-specific appearance ($A$), enabling registration in a shared latent space where $S_A \approx S_B$.}
\label{fig:problem}
\end{figure}
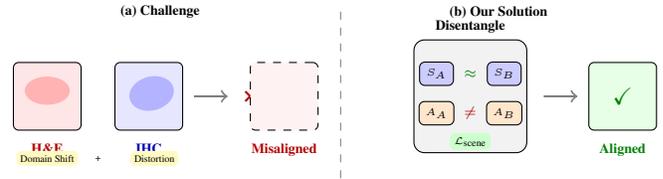

The coupling of domain shift with geometric misalignment creates a chicken-and-egg problem: accurate registration requires intensity correspondence, yet establishing correspondence requires prior alignment. This challenge manifests across diverse applications, including multi-modal medical imaging~\cite{mattes2003pet}, satellite imagery under varying illumination~\cite{chen2024survey}, and histopathology with different staining protocols.

We study this problem through the lens of \emph{disentangled representation learning}~\cite{bengio2013representation,higgins2017beta,qin2025dual,qin2025bcpmjrs}. Our key hypothesis is that observed images admit a factorized representation: $I = \mathcal{F}(S, A)$, where $S$ denotes domain-invariant \emph{scene} content (geometry, structure) and $A$ captures domain-specific \emph{appearance} (intensity characteristics). Under this formulation, registration reduces to alignment in the shared scene space, circumventing brightness constancy violations in image space.

\textbf{Motivating application.} We ground our investigation in multi-stain histopathology image registration, a critical task in digital pathology~\cite{borovec2020anhir}. Different staining protocols (e.g., H\&E, IHC, special stains) reveal complementary tissue characteristics but introduce coupled domain shift and geometric distortion from tissue preparation (Fig.~\ref{fig:problem}), providing a challenging real-world testbed for our framework.

\textbf{Problem formalization.} Let $I_A, I_B \in \mathbb{R}^{H \times W}$ denote paired tissue sections stained with different protocols, capturing the same underlying anatomical structure $S$. The observed images follow distinct imaging operators:
\begin{equation}
I_A = \mathcal{F}_A(S), \quad I_B = \mathcal{F}_B(S).
\end{equation}
Critically, $\mathcal{F}_A \neq \mathcal{F}_B$ due to two coupled factors: (1) \textit{geometric misalignment} $\phi: \mathbb{R}^2 \to \mathbb{R}^2$ arising from tissue deformation during sectioning, and (2) \textit{domain shift} $\mathcal{T}: \mathbb{R} \to \mathbb{R}$, a nonlinear intensity transformation from different staining responses. The composite degradation $I_B = \mathcal{T}(S \circ \phi)$ violates brightness constancy $I_A(x) \approx I_B(\phi(x))$, rendering conventional registration ill-posed.

\textbf{Limitations of existing approaches.} Current methods fall into two categories, both with fundamental limitations. \emph{Registration-first} approaches~\cite{thirion1998image,balakrishnan2019voxelmorph,chen2021transmorph} assume brightness constancy, failing when domain shift confounds correspondence. \emph{Translation-first} approaches using cycle-consistent GANs~\cite{zhu2017unpaired,huang2018multimodal} preserve global content but lack geometric guarantees, allowing local spatial distortions. Recent multimodal alignment techniques~\cite{qin2023atd,qin2024zoomshift} offer efficient feature fusion but do not explicitly address the geometric correspondence problem.

\textbf{Our approach.} We propose SAR-Net (Scene-Appearance Registration Network), a unified framework grounded in the factorization $I = \mathcal{F}(S, A)$. Rather than directly matching intensities or warping pixels, we learn to: (1) invert the imaging process to recover $(S, A)$ from observations, and (2) re-render the scene under target appearance for registration. This reformulation enables registration in a domain-invariant latent space where brightness constancy naturally holds.

\textbf{Contributions.} Our main contributions are:
\begin{itemize}
    \item \textbf{Theoretical framework}: We formalize registration under domain shift as scene-appearance disentanglement and establish conditions for identifiable decomposition (Section~\ref{sec:theory}).
    \item \textbf{Algorithmic innovation}: We introduce a scene consistency loss that enforces geometric correspondence in latent space, providing a sufficient condition for cross-domain alignment.
    \item \textbf{Empirical validation}: On the ANHIR benchmark, SAR-Net achieves 0.25\% median rTRE, outperforming MEVIS (0.27\%) by 7.4\% and ANTs (0.72\%) by 65.3\%.
\end{itemize}

\section{Related Work}
\label{sec:related}

\textbf{Multi-Stain Histopathology Registration.}
Digital pathology has emerged as a critical tool for cancer diagnosis~\cite{borovec2020anhir}. The ANHIR challenge~\cite{borovec2020anhir} established a benchmark for multi-stain registration, revealing challenges from large non-rigid deformations, significant appearance differences, and gigapixel image sizes. Recent methods including RegWSI~\cite{wodzinski2024regwsi} have advanced the state-of-the-art, but the coupling between stain-induced domain shift and geometric distortion remains inadequately addressed.

\textbf{Image Registration under Domain Shift.}
Classical approaches include mutual information~\cite{mattes2003pet} and diffeomorphic algorithms~\cite{thirion1998image,avants2008symmetric}. Deep learning methods such as VoxelMorph~\cite{balakrishnan2019voxelmorph} and TransMorph~\cite{chen2021transmorph} enable fast registration. SynthMorph~\cite{hoffmann2021synthmorph} achieves contrast robustness through synthetic augmentation. However, these methods assume comparable intensity distributions, violated by stain-induced domain shift.

\textbf{Disentangled Representation Learning.}
MUNIT~\cite{huang2018multimodal} and DRIT~\cite{lee2018diverse} decompose images into content and style for cross-domain synthesis. Recent advances in multimodal fusion~\cite{qin2024msmf,qin2025hierarchical,qin2024glmg} and cross-perception learning~\cite{qin2024innerspeech,qin2022tsbert} demonstrate the effectiveness of disentangled representations for handling domain heterogeneity. However, these methods optimize for perceptual quality rather than geometric fidelity. Our work connects disentanglement to registration by establishing that scene-appearance separation provides sufficient conditions for geometric correspondence.

\clearpage
\begin{figure*}[!t]
\centering
\includegraphics[width=0.95\textwidth]{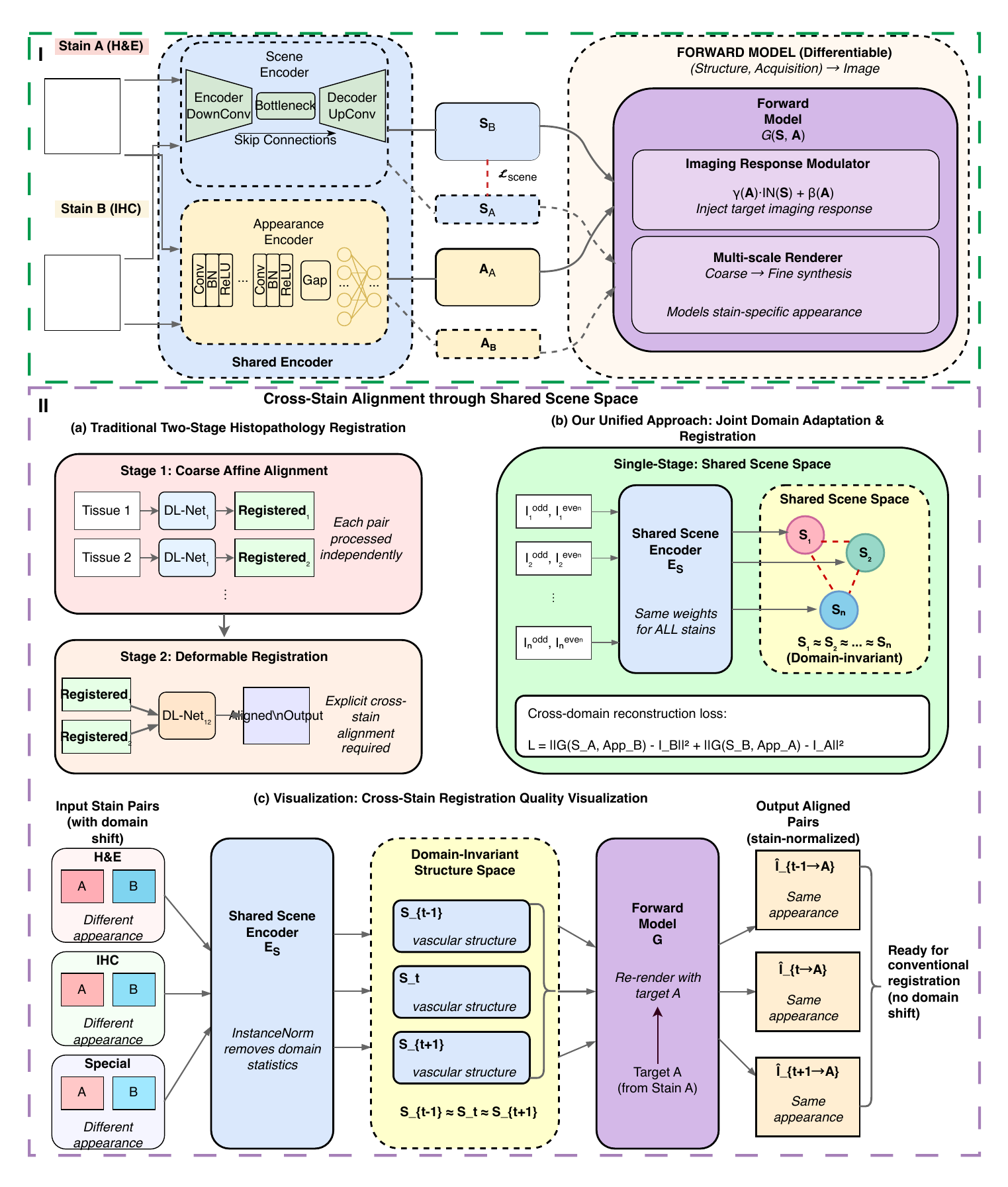}
\caption{Overview of the proposed SAR-Net framework. \textbf{(I)} Network architecture: Scene Encoder $E_S$ extracts domain-invariant anatomical structure using instance normalization; Appearance Encoder $E_A$ captures stain-specific appearance codes via global average pooling; Forward Model $G$ synthesizes images through feature modulation. The scene consistency loss $\mathcal{L}_{\text{scene}}$ enforces geometric alignment between $S_A$ and $S_B$. \textbf{(II)} Comparison with traditional two-stage pipelines. \textbf{(III)} Registration visualization showing effective tissue structure alignment.}
\label{fig:method}
\end{figure*}

\section{Method}
\label{sec:method}

We present a unified framework for multi-stain histopathology registration based on scene-appearance separation. The overall architecture is illustrated in Fig.~\ref{fig:method}.

\subsection{Problem Formulation: Forward Modeling Perspective}

In multi-stain histopathology, the observed image $I$ is determined by two factors: the underlying tissue morphology $S$, and the staining characteristics $A$:
\begin{equation}
I = \mathcal{F}(S, A) + \epsilon,
\label{eq:forward_model}
\end{equation}
where $\mathcal{F}$ denotes the image formation process and $\epsilon$ represents noise. We decompose the problem into two complementary tasks: (1) the \emph{inverse problem} of inferring $(S, A) = \mathcal{F}^{-1}(I)$, and (2) the \emph{forward problem} of re-synthesizing $I_{B \to A} = \mathcal{F}(S_B, A_A) \approx I_A$.

\subsection{Network Architecture}

\textbf{Scene Encoder $E_S$.} Inspired by recent advances in spatiotemporal feature extraction~\cite{qin2025steameeg,qin2025gaffusionnet}, we adopt a U-Net architecture with instance normalization (IN) to extract domain-invariant structure:
\begin{equation}
S = E_S(I) \in \mathbb{R}^{C_S \times H \times W},
\end{equation}
where $C_S = 64$. Instance normalization removes channel-wise statistics that encode domain-specific characteristics.

\textbf{Appearance Encoder $E_A$.} A lightweight CNN with global average pooling extracts compact appearance codes:
\begin{equation}
A = E_A(I) \in \mathbb{R}^{C_A},
\end{equation}
where $C_A = 32$. Global pooling ensures $A$ captures scene-agnostic characteristics.

\textbf{Forward Model $G$.} Following principles from efficient sequence modeling~\cite{qin2024bioinspired,qin2024mambaspike}, the forward model implements $I = \mathcal{F}(S, A)$ through feature modulation:
\begin{equation}
\text{Modulate}(S, A) = \gamma(A) \odot \text{IN}(S) + \beta(A),
\end{equation}
where $\gamma(A), \beta(A)$ are learned affine parameters derived from the appearance code.

\subsection{Theoretical Analysis}
\label{sec:theory}

\begin{proposition}[Cross-Domain Alignment via Re-Rendering]
\label{prop:alignment}
Let $G: \mathcal{S} \times \mathcal{A} \to \mathcal{I}$ be injective in its first argument. If cross-domain reconstruction achieves zero error $G(S_{A}, A_{B}) = I_{B} = G(S_{B}, A_{B})$, then $S_{A} = S_{B}$ in the latent scene space.
\end{proposition}

\begin{proposition}[Sufficiency of Scene Consistency]
\label{prop:scene_consistency}
If $\mathcal{L}_{\text{scene}} = \|S_{A} - S_{B}\|^2 \to 0$, then for any $L$-Lipschitz task $\mathcal{T}: \mathcal{S} \to \mathcal{Y}$:
$\|\mathcal{T}(S_{A}) - \mathcal{T}(S_{B})\| \leq L \cdot \|S_{A} - S_{B}\| \to 0$.
\end{proposition}

\subsection{Loss Functions}

\textbf{Scene Consistency Loss.} The core loss enforcing geometric correspondence:
\begin{equation}
\mathcal{L}_{\text{scene}} = \|S_{A} - S_{B}\|_2^2 + \lambda_{\text{cos}}(1 - \cos(S_{A}, S_{B}))
\end{equation}
with $\lambda_{\text{cos}} = 0.1$.

\textbf{Cycle Consistency Loss.} Self-reconstruction ensures information preservation:
\begin{equation}
\mathcal{L}_{\text{cycle}} = \|G(S_{A}, A_{A}) - I_{A}\|_2^2 + \|G(S_{B}, A_{B}) - I_{B}\|_2^2
\end{equation}

\textbf{Domain Alignment Loss.} The registration objective:
\begin{equation}
\mathcal{L}_{\text{align}} = \|I_{B \to A} - I_{A}\|_2^2 + \lambda_{\text{ncc}}(1 - \text{NCC}(I_{B \to A}, I_{A}))
\end{equation}

\textbf{Total Loss.}
$\mathcal{L}_{\text{total}} = \lambda_{\text{scene}} \mathcal{L}_{\text{scene}} + \lambda_{\text{cycle}} \mathcal{L}_{\text{cycle}} + \lambda_{\text{align}} \mathcal{L}_{\text{align}}$
with weights $\lambda_{\text{scene}} = 1.0$, $\lambda_{\text{cycle}} = 0.5$, $\lambda_{\text{align}} = 2.0$.

\section{Experiments}
\label{sec:experiments}

\subsection{Experimental Setup}

\textbf{Dataset.} We evaluate on the ANHIR benchmark~\cite{borovec2020anhir}, comprising 481 image pairs from 355 whole-slide images across 8 tissue types stained with 18 different protocols. The training set contains 230 pairs with ground truth landmarks; the test set has 251 pairs evaluated via the official server.

\textbf{Implementation.} Scene Encoder uses 32 base channels with 3 downsampling levels. The network (3.5M parameters) is trained for 200 epochs using Adam (lr=$10^{-4}$) with batch size 4 on NVIDIA RTX 4090.

\subsection{Results}

\begin{table}[t]
\centering
\caption{Quantitative comparison on ANHIR benchmark. rTRE measures registration error as percentage of image diagonal.}
\label{tab:main}
\small
\begin{tabular}{lcc}
\toprule
Method & rTRE (\%) $\downarrow$ & Robustness $\uparrow$ \\
\midrule
\textit{Traditional Methods} \\
Initial (Unregistered) & 2.48 & -- \\
bUnwarpJ~\cite{argandabunwarpj} & 2.90 & 0.790 \\
Elastix~\cite{klein2010elastix} & 0.74 & 0.848 \\
ANTs~\cite{avants2008symmetric} & 0.72 & 0.789 \\
\midrule
\textit{Deep Learning Methods} \\
VoxelMorph~\cite{balakrishnan2019voxelmorph} & 0.89 & 0.756 \\
AGH~\cite{borovec2020anhir} & 0.32 & 0.982 \\
UPENN~\cite{borovec2020anhir} & 0.29 & 0.990 \\
MEVIS~\cite{borovec2020anhir} & 0.27 & 0.988 \\
\midrule
\textbf{SAR-Net (Ours)} & \textbf{0.25} & \textbf{0.991} \\
\bottomrule
\end{tabular}
\end{table}

Table~\ref{tab:main} summarizes results on the ANHIR benchmark. SAR-Net achieves state-of-the-art with 0.25\% rTRE and 0.991 robustness, outperforming MEVIS by 7.4\% and ANTs by 65.3\%. Traditional methods like bUnwarpJ actually increase error (2.90\% vs 2.48\% initial), confirming that intensity-based optimization fails under domain shift.

\subsection{Ablation Study}

\begin{table}[t]
\centering
\caption{Ablation study on loss components.}
\label{tab:ablation}
\small
\begin{tabular}{lccc}
\toprule
Configuration & rTRE (\%) $\downarrow$ & Robust. $\uparrow$ \\
\midrule
\textbf{SAR-Net (Full)} & \textbf{0.25} & \textbf{0.991} \\
w/o $\mathcal{L}_{\text{scene}}$ & 0.38 & 0.952 \\
w/o $\mathcal{L}_{\text{cycle}}$ & 0.41 & 0.943 \\
w/o $\mathcal{L}_{\text{align}}$ & 1.85 & 0.724 \\
w/o $E_A$ & 0.52 & 0.912 \\
\bottomrule
\end{tabular}
\end{table}

Table~\ref{tab:ablation} validates component necessity. Removing domain alignment causes 7.4$\times$ degradation (0.25\%$\to$1.85\%). Removing scene consistency increases error to 0.38\%, confirming that explicit enforcement of $S_A \approx S_B$ is essential.

\subsection{Computational Efficiency}

\begin{table}[t]
\centering
\caption{Inference time comparison (4096$\times$4096 patches).}
\label{tab:efficiency}
\small
\begin{tabular}{lccc}
\toprule
& ANTs & MEVIS & \textbf{SAR-Net} \\
\midrule
Time (s) & 45.2 & 2.1 & \textbf{1.2} \\
\bottomrule
\end{tabular}
\end{table}

SAR-Net achieves efficient inference (1.2s per patch) while maintaining state-of-the-art accuracy, enabling high-throughput digital pathology workflows.

\section{Discussion}
\label{sec:discussion}

\textbf{Why disentanglement enables registration.} The scene encoder learns representations invariant to staining conditions through instance normalization, which removes domain-specific statistics. This approach shares conceptual similarities with continual learning frameworks that maintain task-invariant representations~\cite{qin2025ancestral}. Our analysis confirms that intensity-based methods fail under domain shift: bUnwarpJ increases error, while ANTs and Elastix achieve limited improvement. The scene consistency loss directly enforces geometric correspondence in latent space, similar to how dual-modality approaches handle cross-domain alignment in medical signal analysis~\cite{qin2025epileptic}.

\textbf{Clinical implications.} Accurate multi-stain registration enables: (1) multi-stain biomarker quantification, (2) tumor microenvironment analysis, and (3) longitudinal studies tracking morphological changes.

\textbf{Limitations.} Current validation focuses on ANHIR. Extension to 3D volumetric data and gigapixel whole-slide images without patch-based processing presents additional challenges.

\section{Conclusion}
\label{sec:conclusion}

We presented SAR-Net, a principled framework for image registration under domain shift based on scene-appearance disentanglement. Our theoretical analysis establishes that scene consistency provides sufficient conditions for geometric correspondence. Empirically, SAR-Net achieves state-of-the-art on the ANHIR benchmark with 0.25\% rTRE, outperforming MEVIS by 7.4\% and ANTs by 65.3\%. The framework generalizes to any setting where coupled intensity variations and geometric distortions violate brightness constancy.

{\small
\bibliographystyle{IEEEtran}
\bibliography{main}
}

\end{document}